\documentclass[10pt,twocolumn,letterpaper]{article}

\usepackage{iccv}
\usepackage{times}
\usepackage{epsfig}
\usepackage{graphicx,floatrow}
\usepackage{subfig}
\usepackage{multirow}
\usepackage{amsmath}
\usepackage{amssymb}
\usepackage{latexsym}
\usepackage[lined,boxed,commentsnumbered, ruled]{algorithm2e}
\usepackage[pagebackref=true,breaklinks=true,letterpaper=true,colorlinks,bookmarks=false]{hyperref}

\iccvfinalcopy % *** Uncomment this line for the final submission

\ificcvfinal\fi

\begin{document}
% \vspace{-10em}
%%%%%%%%% TITLE
\title{Zero-Shot Learning by Generating Pseudo Feature Representations}
\author{Jiang Lu\textsuperscript{$\ast,\dagger,\ddagger$}\quad Jin Li\textsuperscript{$\ast,\dagger$}\quad Ziang Yan\textsuperscript{$\dagger$}\quad Changshui Zhang\textsuperscript{$\dagger$}\\
\textsuperscript{$\dagger$}Department of Automation, Tsinghua University\\
State Key Lab of Intelligence Technologies and Systems\\ 
Tsinghua National Laboratory for Information Science and Technology (TNList), Beijing, China
\\\quad 
\textsuperscript{$\ddagger$}China Marine Development and Research Center (CMDRC), Beijing, China\\
% Beijing, China\\
{\tt\small \{lu-j13,lijin14,yza15\}@mails.tsinghua.edu.cn, zcs@mail.tsinghua.edu.cn}}
\maketitle
\renewcommand{\thefootnote}{\fnsymbol{footnote}}
\footnotetext[1]{Denotes equal contribution}
\renewcommand{\thefootnote}{\arabic{footnote}}

%\thispagestyle{empty}

%%%%%%%%% ABSTRACT
\begin{abstract}
	Zero-shot learning (ZSL) is a challenging task aiming at recognizing novel classes without any training instances. In this paper we present a simple but high-performance ZSL approach by generating pseudo feature representations (GPFR). Given the dataset of seen classes and side information of unseen classes (\eg attributes), we synthesize feature-level pseudo representations for novel concepts, which allows us access to the formulation of unseen class predictor. Firstly we design a Joint Attribute Feature Extractor (JAFE) to acquire understandings about attributes, then construct a cognitive repository of attributes filtered by confidence margins, and finally generate pseudo feature representations using a probability based sampling strategy to facilitate subsequent training process of class predictor. We demonstrate the effectiveness in ZSL settings and the extensibility in supervised recognition scenario of our method on a synthetic colored MNIST dataset (C-MNIST). For several popular ZSL benchmark datasets, our approach also shows compelling results on zero-shot recognition task, especially leading to tremendous improvement to state-of-the-art mAP on zero-shot retrieval task. 
\end{abstract}
\vspace{-1em}
%%%%%%%%% BODY TEXT
\section{Introduction}\label{chap-introduction}
Although large scale classification based on supervised learning has achieved major successes in recent years 
by deep learning \cite{russakovsky2015imagenet,krizhevsky2012imagenet,simonyan2014very,szegedy2015going,he2016deep}, the collection and annotation of huge amounts of training data for each class become a bottleneck for many visual recognition tasks. The increasing new categories and few available training examples are forcing us to develop more efficient learning paradigms.
\begin{figure}[t]
\begin{center}
\includegraphics[width=1.0\linewidth]{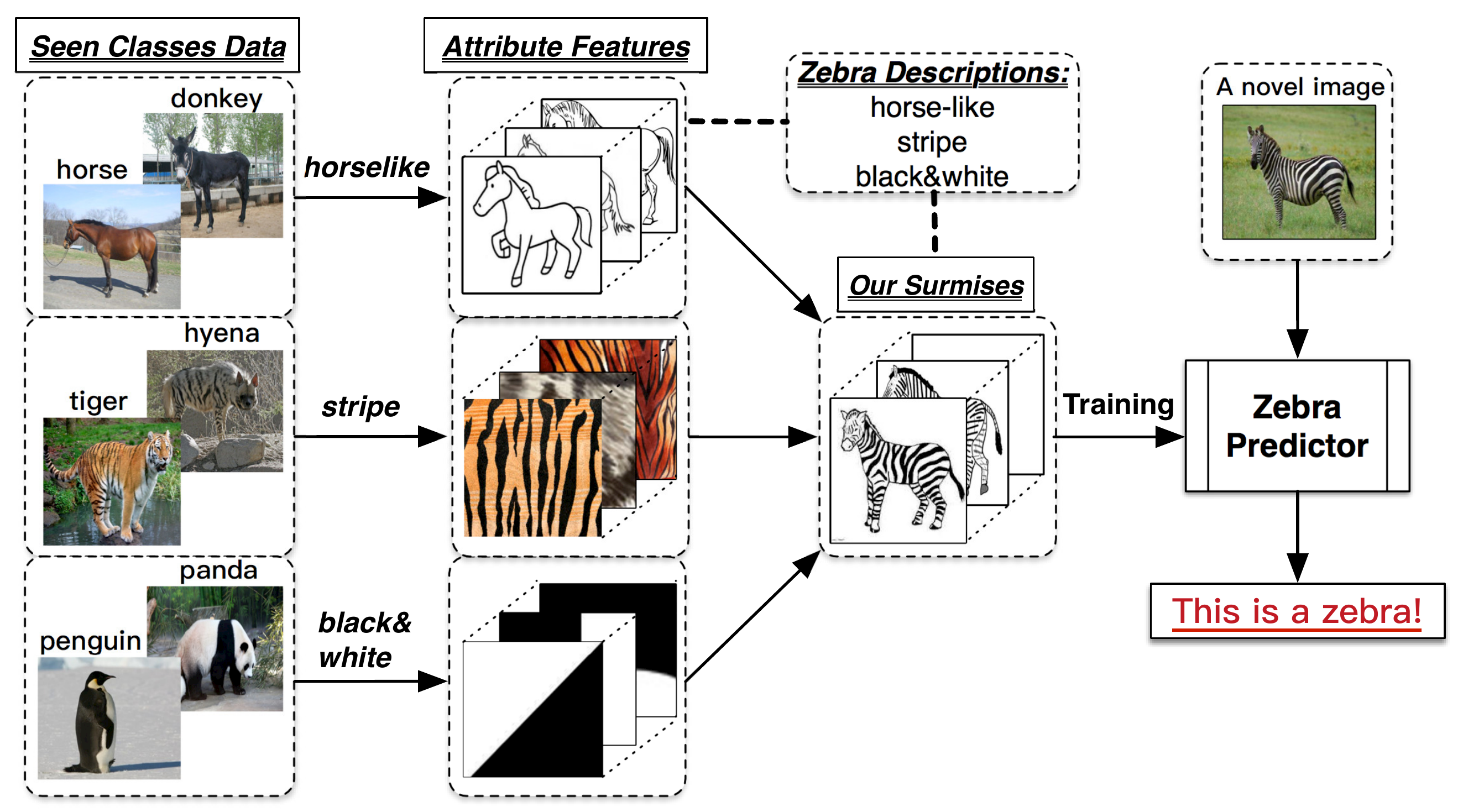}
\end{center}
\vspace{-1.5em}
   \caption{Illustration of our intuition for ZSL tasks. Firstly, we capture some understandings from prepared dataset about what features represent \emph{horselike/stripe/black\&white} respectively. Then, these attribute features associated with key descriptions of zebra will be combined into lots of synthetic surmises about zebras' appearance, which makes the training of zebra predictor possible. Finally, a novel zebra image can be inferred by our well-trained predictor.}
\label{fig:idea}
\end{figure}
\vspace{0.1em}

Zero-shot learning (ZSL) aims to recognize previously unseen classes without labelled training instances, which has gained growing attention recently \cite{lampert2009learning,akata2013label,farhadi2009describing,palatucci2009zero}. For ZSL tasks, some intermediate semantic properties, such as attributes \cite{farhadi2009describing,lampert2009learning} or category hierarchies \cite{rohrbach2011evaluating}, are usually revealed and shared for seen and unseen classes, acting as side information by which the unseen classes could be inferred rationally. Explicitly, in ZSL settings, the dataset of seen classes is well labelled with categories and attributes tags, whereas the unseen classes are faced with absence of training instances but presence of their attribute descriptions. The purpose of ZSL for visual recognition is to predict for each novel image which of unseen classes it belongs to.

Previous works for ZSL can be broadly divided into two major categories. Some works advised to perform a two-stage probability based method, namely attribute prediction and then classification inference \cite{farhadi2009describing,lampert2014attribute,yu2010attribute,suzuki2014transfer}, while others attempted to decompose the ZSL tasks into two sub-tasks, semantic embedding and similarity measurement, whose performance is excessively reliant on the capability of shared semantic embedding spaces \cite{akata2013label,akata2015evaluation,romera2015embarrassingly,zhang2015zero,zhang2016zero,bucher2016improving}.
 \begin{figure*}[tbp]
\begin{center}
\includegraphics[width=1\linewidth]{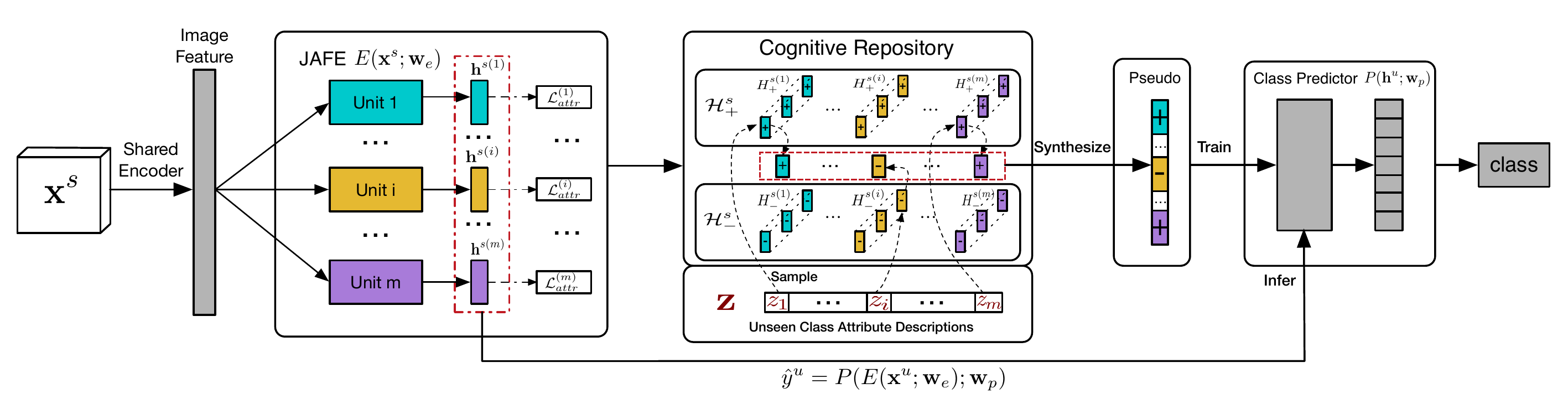}
\end{center}
\vspace{-2.0em}
  \caption{Framework of our GPFR model. The bars in different colors denote respectively different attribute feature vectors. The bars with ``+'' are sufficiently positive feature vectors on the presence of its corresponding attribute, and it is opposite for the ones with ``-''. Two parts of GPFR need to be trained from scarcity, JAFE and class predictor. Cognitive repository can be constructed based on the well-trained JAFE. Pseudo representations of unseen classes, which
  guide the formation of class predictor, are generated via a probability based sampling. Best viewed in color.}
\label{fig:framework}
\end{figure*}

In this paper we present a novel approach for ZSL via generating pseudo feature representations, called GPFR, inspired by humans' behaviors of recognizing a novelty. It is generally known that the process to recognize novel concepts for most of people is abstractively from individuality to generality, and then from generality to individuality. More specifically, assuming the fundamental understandings about what features represent the attributes \emph{horselike/stripe/black\&white} are obtained from some prepared images, like \emph{horse, donkey, tiger, hyena, penguin, panda} etc, one can surmise roughly what a zebra looks like if told zebra has above attributes. As illustrated in Fig.~\ref{fig:idea}, our intuition is to firstly learn some credible feature representations for each attribute by utilizing prepared dataset of seen classes, and then summarize these attribute representations into an combined representation, according to the specified attribute descriptions of each unseen class. Finally, the combined representation can be viewed as a so-called pseudo representation of unseen class image, which will offer valuable guidance for the training of unseen class predictor.

% \vspace{-0.em}
In our ZSL framework, leveraging the Convolutional Neural Network (CNN) based image features \cite{simonyan2014very}, we firstly train a Joint Attribute Feature Extractor (JAFE) in which each fundamental unit is put in charge of the extraction of one attribute feature vector. Regardless of labels of seen classes, we extract all possible feature vectors for every attribute tag by JAFE, then assembled these vectors into a cognitive repository of attributes based on a confidence margin filter. According to the attribute descriptions of unseen classes, a probability based sampling strategy is exploited to select some attribute feature vectors from this cognitive repository to synthesize combined vectors. This strategy allows us access to lots of synthetic feature vectors for a specified unseen class, called pseudo feature representations, which fills the gaps between training domain and test domain and achieves data augmentation in feature level as well for supervised recognition from another perspective. Taking these pseudo feature representations as inputs, a multi-classes predictor for unseen classes can be learned. During test time, a novel image goes through the JAFE to generate its combined feature vector over all attributes, followed by above well-trained predictor to perform the ultimate inference. Results on a synthetic colored MNIST dataset (C-MNIST) demonstrate the effectiveness and extensibility of our GPFR. Furthermore, its performance on several ZSL benchmark datasets show improvements to state-of-the-art results, especially for zero-shot retrieval mAP (\ie mean average precision).
\vspace{0.2em}

We conclude our contributions as follows. First, we develop a concise GPFR framework for ZSL tasks, which fills the gaps between seen and unseen concepts and achieves compelling results on four challenging ZSL benchmark datasets. Second, our GPFR model realizes inherently the data augmentation in feature level, which can be easily extended to supervised learning scenario. 
\vspace{-0.5em}
%---------------------------------------------------------------------------
\section{Methodology}\label{chap-method}
%-------------------------------------------------------------------------
\subsection{Overview} \label{chap-method-overview}
\paragraph{Basic formulation.} Suppose we represent each image by a $d$-dimensional visual feature vector 
$\mathbf{x}$ and represent each attribute description by an $m$-dimensional binary vector $\mathbf{a}$ where $m$ is the number of attributes. Let $c$ represent the number of unseen classes, and let $T$ represent the label set of unseen classes, \ie $|T|=c$. Moreover, let $\mathbf{S}=\left\{(\mathbf{x}^{s}_{i},\mathit{y}^{s}_{i},\mathbf{a}_{i})\right\}^{N_{s}}_{i=1}$ represent the labelled dataset of seen classes with $\mathit{N_{s}}$ images, and $\mathbf{U}=\{(\mathbf{x}^{u}_{i},\mathit{y}^{u}_{i})\}^{N_{u}}_{i=1}$ represent the unlabelled dataset of unseen classes with $\mathit{N_{u}}$ images, where $y$ denote the class label. Besides, the prepared class-level attribute descriptions for all unseen classes are preprocessed into the probabilistic form where its values represent the possibilities of presence for attributes, which are denoted by $\mathbf{Z} = \left\{\mathbf{z}^t, t\in T \right\}$, where $\mathbf{z}^t\in \mathbb{R}^m$ denotes the attribute descriptions of unseen class $t$. In this way our goal for ZSL is to utilize the prepared $\mathbf{S}$ and $\mathbf{Z}$ to predict $\mathit{y}^u$ for unseen class image $\mathbf{x}^u$. Concerning on our method, let $E\left(\mathbf{x};\mathbf{w}_{e}\right)$ be a function parameterized by $\mathbf{w}_e$ which maps $\mathbf{x}$ to a combined hidden representation $\mathbf{h}$ concatenated by $m$ sub-vectors $\mathbf{h}^{(i)}, i=1,\ldots,m$, where $\mathbf{h}^{(i)}$ denote the $i$-th sub-vector of $\mathbf{h}$. For clarity, we denote by $\mathbf{h}^s$ the combined hidden representation of seen classes, and  $\mathbf{h}^u$ the pseudo hidden representation of unseen classes. Thus $\mathbf{h}^{s(i)}$ and $\mathbf{h}^{u(i)}$ denote respectively the $i$-th sub-vector of $\mathbf{h}^s$ and $\mathbf{h}^u$.
Let $P\left(\mathbf{h};\mathbf{w}_p\right)$ be a classification function parameterized by $\mathbf{w}_p$ which maps a combined hidden representation $\mathbf{h}$ to the unseen class prediction $\hat{y}$. It's remarkable that function $E$ is learned via dataset $\mathbf{S}$ but $P$ via synthetic pseudo feature representations of unseen classes.
\vspace{0.25em}

Like previous works \cite{zhang2015zero,zhang2016zero,bucher2016improving}, we utilize pre-trained CNN model \cite{simonyan2014very} to obtain image features. Our GPFR model is depicted in Fig.~\ref{fig:framework}. The JAFE, \ie $E\left(\mathbf{x};\mathbf{w}_{e}\right)$, aims to extract all attribute feature vectors and then form a combined feature representation. The class predictor, \ie $P\left(\mathbf{h};\mathbf{w}_p\right)$, is adjustable with specified tasks, such as prediction on seen classes (\ie fully supervised recognition) instead of unseen classes. Based on dataset $\mathbf{S}$, we adopt a joint training strategy for JAFE to reduce time cost. Due to absence of training images for unseen classes, we employ a confidence margin based filter to construct the cognitive repository for all attributes. Then, we regard attribute descriptions of unseen classes as sampling probabilities to generate pseudo feature representations from cognitive repository, which allows us access to the training of unseen class predictor.
% \vspace{-0.5em}
%-------------------------------------------------------------------------
\subsection{Joint Attribute Feature Extractor (JAFE)} \label{chap-method-extraction}
In this section we detail our JAFE framework. Given the labelled dataset $\mathbf{S}$, we can train our JAFE regardless of their class-level labels. In our paper, we construct our JAFE by single-layer perceptron (SLP) or multi-layer perceptron (MLP) group, which of course can be replaced by other feature extraction model. Explicitly, each unit in JAFE is a SLP or MLP, assigned to deal with the extraction of one specified attribute feature vector, \ie $\mathbf{h}^{(i)}$ as notated in Sec.~\ref{chap-method-overview}. Then the combined feature representation is defined as follows:
\begin{equation}
\mathbf{h}=\mathcal{C}(\mathbf{h}^{(1)},\ldots,\mathbf{h}^{(m)})=E\left(\mathbf{x};\mathbf{w}_e\right),
\label{eq:1}
\end{equation}
where $\mathcal{C}$ denotes concatenation operation on sub-vectors. Of course, we can set different feature dimensions for different attributes, but here we let all $\mathbf{h}^{(i)}$, $i=1,\ldots,m$, are $l$-dimensional vectors for simplicity, thus $\mathbf{h}\in \mathbb{R}^{ml}$.
\vspace{-1.15em}
\paragraph{Joint training framework.}
Given a seen class image $\mathbf{x}^s$ with its attribute description $\mathbf{a}$ which has been processed into a binary vector by us prior to training, we utilize the 2-way Softmax classifier as our attribute predictor. We represent the probability that $\mathbf{x}^s$ is positive about the $i$-th attribute by $\mathit{p}_i(\mathbf{h}^{s(i)})$, which is predicted by the attribute predictor connected with the $i$-th unit of JAFE. We minimize the negative binomial cross-entropy for the $i$-th attribute prediction:
\begin{equation}
\mathcal{L}_{attr}^{(i)}=a^{(i)} \log \mathit{p}_i(\mathbf{h}^{s(i)})+(1-a^{(i)})\log (1-\mathit{p}_i(\mathbf{h}^{s(i)})),
\label{eq:2}
\end{equation}
where $a^{(i)}\in \{0,1\}$ is ground truth label of the $i$-th attribute tag in $\mathbf{a}$. Since $\mathbf{h}^{s(i)}$ is obtained from $E\left( \mathbf{x}^s; \mathbf{w}_e\right)$, as showed in Eq.~\ref{eq:1}, we can optimize our parameter $\mathbf{w}_e$ by designing the following joint loss function:
\begin{equation}
\mathcal{L}_{joint}=\sum_{i=1}^{m}\alpha_i\mathcal{L}_{attr}^{(i)},
\label{eq:3}
\end{equation} 
where $\alpha_{i}$, $i=1,\ldots,m$, achieve a weight assignment between all attribute predictions. Especially, we formulate our training loss for JAFE and attribute predictors in one-sample case, as showed in Eq.~\ref{eq:2} and \ref{eq:3}, but we can extend it to a mini-batch version just by an average operation, which can be optimized by gradient back-propagation algorithm.

%-----------------------------------------------------------------------------
\subsection{Generating Pseudo Feature Representations} \label{chap-method-generating}
For seen class image $\mathbf{x}^s$, the well-trained JAFE and attribute predictors lead to attribute feature vectors $\{\mathbf{h}^{s(i)}\}^m_{i=1}$ associated with their corresponding positive estimations $\{p_i(\mathbf{h}^{s(i)})\}^m_{i=1}$, which lays the foundation for subsequent pseudo feature representations generation.
\vspace{-1.1em}
\paragraph{Cognitive repository of attributes based on confidence margins.} 
Just as humans often summarize the characteristics of a certain attribute from some different things, our approach also follow this inspiration to build a cognitive repository for some credible attribute-level feature vectors. Our intuition tells us that the ambiguous attribute-level feature vectors predicted by attribute predictors may not be applicable for synthesizing pseudo combined representations, since their uncertainty will affect our judgments. In this way we introduce a confidence margin tuple $(\Delta_{+},\Delta_{-})$ to guarantee the truth-reliability of cognitive faculties for attributes, where $\Delta_{+}\in \left[0.5,1\right)$ and $\Delta_{-}\in\left(0,0.5\right]$. Explicitly, based on $\{\mathbf{h}^{s(i)}\}^m_{i=1}$ and $\{p_i(\mathbf{h}^{s(i)})\}^m_{i=1}$, we construct our cognitive repository of attribute feature vectors, denoted by $\mathcal{H}_{+}^s=\{H_{+}^{s(i)}\}^m_{i=1}$ and $\mathcal{H}_{-}^s=\{H_{-}^{s(i)}\}^m_{i=1}$, which represent respectively the set of positive and negative attribute-level feature vector subsets about all attributes. The cognitive repository $H_{+}^{s(i)}$ and $H_{-}^{s(i)}$ are given by: 
\begin{equation}
\begin{aligned}
H_{+}^{s(i)}=\{\mathbf{h}_k^{s(i)},k\in\{j:p_i(\mathbf{h}_j^{s(i)})\geq\Delta_{+}, a_j^{(i)}=1,\\j=1,\ldots,N_s\}\},
\label{eq:4}
\end{aligned}
\end{equation}
\vspace{-0.5em}
\begin{equation}
\begin{aligned}
H_{-}^{s(i)}=\{\mathbf{h}_k^{s(i)},k\in\{j:p_i(\mathbf{h}_j^{s(i)})\leq \Delta_{-}, a_j^{(i)}=0,\\j=1,\ldots,N_s\}\},
 \label{eq:5}
 \end{aligned}
\end{equation}
where $\mathbf{h}_j^{s(i)}$ is the $i$-th attribute feature vector of the $j$-th image in dataset $\mathbf{S}$ extracted by JAFE, and $a^{(i)}_j$ is ground truth label of the $i$-th attribute tag of the $j$-th image. For example, if we set $\Delta_{+}=0.7$ and $\Delta_{-}=0.2$, the above operation means that for $i$-th attribute we just take into account the positive vectors whose positive probability exceeds 0.7 and the negative vectors whose negative probability exceeds 0.8.
\vspace{-2.4em}
\paragraph{Pseudo feature representations generation.} 
As described in Sec.~\ref{chap-method-overview}, the class-level attribute description $\mathbf{Z} = \left\{\mathbf{z}^t, t\in T \right\}$ acts as side information for our ZSL task, where each entry has been preprocessed into a probabilistic form to provide direction for pseudo representations generation. In details, the certain entry $z^t_i$ of $\mathbf{z}^t$ in $\mathbf{Z}$ is viewed as the probability of sampling an attribute-level vector from $H^{s(i)}_+$ instead $H^{s(i)}_-$ with respect to the $i$-th attribute for unseen class $t$, where $t \in T$ , which naturally suggests the synthesizing algorithm in Alg.~\ref{algo:1}. Consequently, for arbitrary unseen class $t\in T$, we can obtain a corresponding pseudo feature-level training dataset $H^t$, and these datasets constitute the complete pseudo training dataset $\mathcal{H}^u = \{H^t,t\in T\}$ with supervision information.
\vspace{-0.8em}
\begin{algorithm}  
\caption{Synthesizing pseudo representations}  
% \SetAlgoNoLine
\small \textbf{Input}: $\mathcal{H}_{+}^s=\{H^{s(i)}_+\}^m_{i=1},\mathcal{H}_{-}^s=\{H^{s(i)}_-\}^m_{i=1}$, unseen class label $t\in T$, its class-level attribute description $\mathbf{z}^t \in \mathbb{R}^m$\\
% $unseen class label $u\in L$ and its class-level attribute description $\mathbf{z}^u \in \mathbb{R}^m$}
\textbf{Initialize:} pseudo size $n$, pseudo training dataset $H^t=\{\}$
\BlankLine    
\For{$k=1$ \textbf{to} $n$}{  
    \For{$i = 1$ \textbf{to} $m$}{  
    	Randomly generate $\epsilon \sim \mathbf{U}(0,1)$;\\
    	\textbf{if} $\epsilon \leq z^t_i$, randomly sample $\mathbf{h}_k^{u(i)} \sim H_+^{s(i)}$;\\
    	\textbf{else}, randomly sample $\mathbf{h}_k^{u(i)} \sim H_-^{s(i)}$;\\
    }  
    $\mathbf{h}_k^{u}\leftarrow\mathcal{C}(\mathbf{h}_k^{u(1)},\ldots,\mathbf{h}_k^{u(m)})$;\\
    Add the $(\mathbf{h}_k^{u},t)$ pair into set $H^t$;
}  
\small \textbf{Output}: $H^t=\{(\mathbf{h}_k^{u},t)\}^n_{k=1}$.
\label{algo:1}
\end{algorithm}
\vspace{-0.5em}
%-------------------------------------------------------------------------
\subsection{Training and Validation for Class Predictor} \label{chap-method-inference}
On the basis of pseudo feature-level training dataset $\mathcal{H}^u = \{H^t,t\in T\}$, we can optionally choose one suitable classifier as our unseen class predictor, namely $P(\mathbf{h}^u;\mathbf{w}_p)$ notated in Sec.~\ref{chap-method-overview}. In this paper we use a $c$-way Softmax classifier to make class prediction :
\begin{equation}
\hat{y}^u=P(\mathbf{h}^u;\mathbf{w}_p) = \mathop{\arg\max}_{t\in T} \mathit{Softmax}(\mathbf{h}^u;\mathbf{w}_p),
\label{eq:6}
\end{equation}
where $\mathbf{w}_p$ is the weights of this Softmax classifier. 
\vspace{-1.0em}
\paragraph{Validation strategy.} Due to absence of training images for unseen classes, we use some seen classes as our validation dataset. More clearly, we add randomly $v$ seen classes into the set of ``unseen'' classes, then our GPFR model is to deal with a ZSL task with $c+v$ unseen classes, whose training process can be supervised by its performance on the $v$ seen classes. Once well trained, the weights of predictor for original $c$ unseen classes will be extracted alone to construct a cutdown predictor just upon the $c$ unseen classes.
\vspace{-1.5em}
\paragraph{Training strategy.} Although many combinations of attribute feature vectors can be used to generate different pseudo representations for unseen classes, we employ an iterative generation strategy to train our class predictor. Explicitly, we use the current $\mathcal{H}^u$ to optimize $\mathbf{w}_p$ for few epochs and then generate a new $\mathcal{H}^u$ by Alg.~\ref{algo:1} to play a repeat until the occurrence of the best validation performance, which intends to avoid overfitting on one random pseudo dataset $\mathcal{H}^u$.
\vspace{0.2em}

Finally, as mentioned in Sec.~\ref{chap-method-overview}, we utilize the well-trained JAFE $E$ and classes predictor $P$ in succession to make inference for a novel image $\mathbf{x}^u$ as follows:
\begin{equation}
\hat{y}^u=P(E(\mathbf{x}^u;\mathbf{w}_e);\mathbf{w}_p),
\label{eq:7}
\end{equation}
which can be implemented easily just by concatenating JAFE and class predictor together.
% \vspace{-0.5em}
%------------------------------------------------------------------------
\section{Experiments}\label{chap-experiments}
We test our method on a synthetic colored MNIST dataset named C-MNIST and other four benchmark image datasets for ZSL recognition, \ie aPascal \& aYahoo (aP\&Y) \cite{farhadi2009describing}, Animals with Attributes (AwA) \cite{lampert2009learning}, Caltech-UCSD Birds-200-2011 (CUB-200-2011) \cite{wah2011caltech} and SUN attribute database \cite{patterson2012sun}. Our codes\footnote{Our codes will be released if this paper is accepted.} are written in Python and the neural network model are implemented with Tensorflow \cite{abadi2016tensorflow} embedded into Keras. We run our codes on a NVIDIA TITAN X (Pascal) GPU.
% \vspace{-1em}
%-------------------------------------------------------------------------
\subsection{Colored MNIST Dataset (C-MNIST)} \label{chap-expre-MNIST}
\begin{figure}[tbp]
\begin{center}
\includegraphics[width=0.77\linewidth]{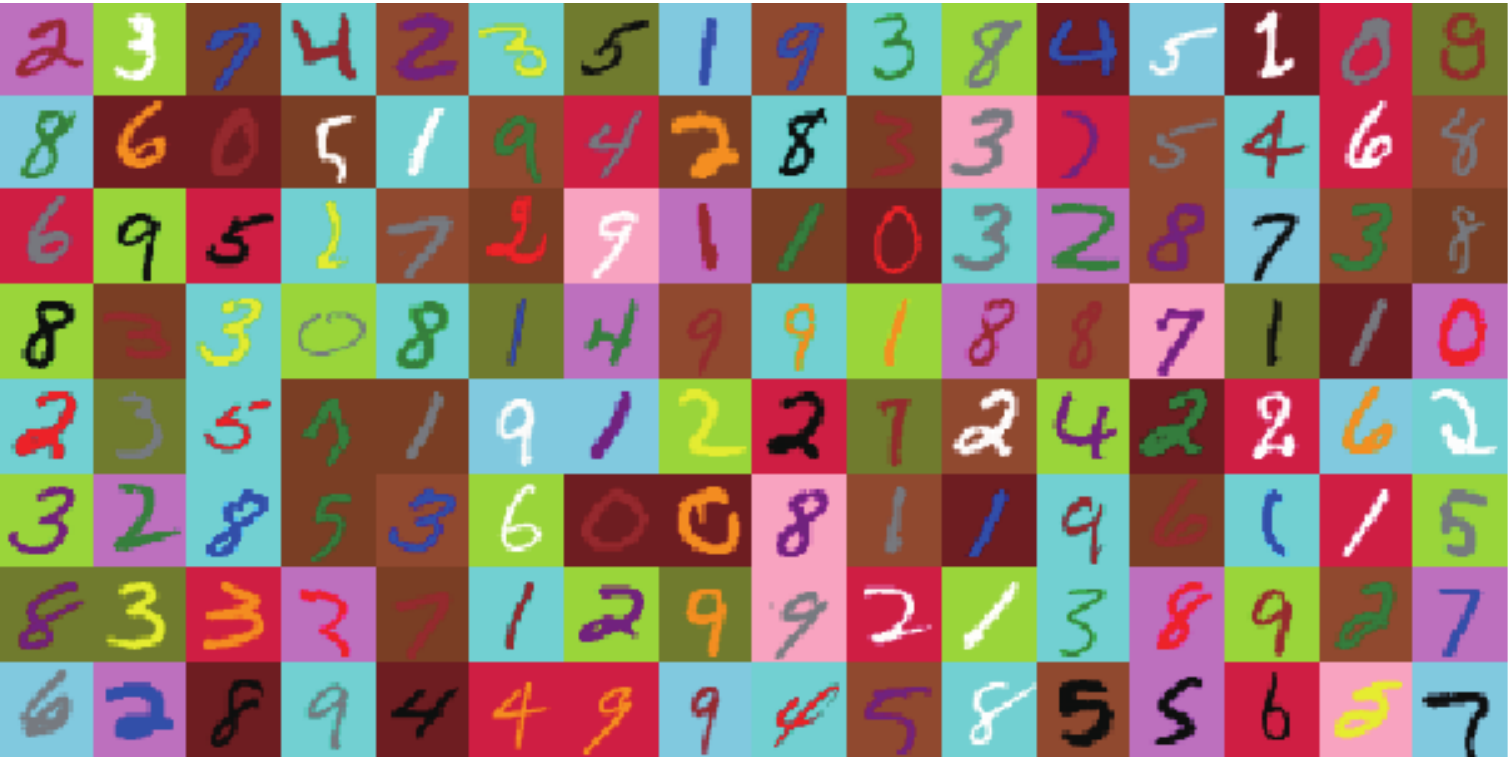}
\end{center}
\vspace{-1.6em}
   \caption{Some examples of C-MNIST. The images from same class own the same digit, b-color and f-color. The size per class is almost 70k/1k=70. Best viewed in color.}
\label{fig:C_MNIST}
\end{figure}

In order to verify the validity and extensibility of our GPFR model, we build a task-specific toy dataset based on MNIST. In details, for original gray images of MNIST, we add randomly 10 different colors into their backgrounds and other 10 different colors into their strokes (also called foregrounds), thus resulting in a new C-MNIST which consists of 70k colored RGB digital images with resolution of $28\times 28$ (60k for training and 10k for testing) from 1k possible combinations (10 digits $\times$ 10 b-colors $\times$ 10 f-colors). Some examples from C-MNIST are showed in Fig.~\ref{fig:C_MNIST}. We regard these 1k combinations as 1k different classes to perform our experiments, thus it essentially can be seen as a synthetic fine-grained recognition dataset. Based on C-MNIST, two types of tasks will be discussed in this section, \ie zero-shot recognition and fully supervised recognition. 
\begin{figure}[tbp]
\centering
\subfloat[Prediction on all 1k classes.]{
 \includegraphics[width=0.48\linewidth]{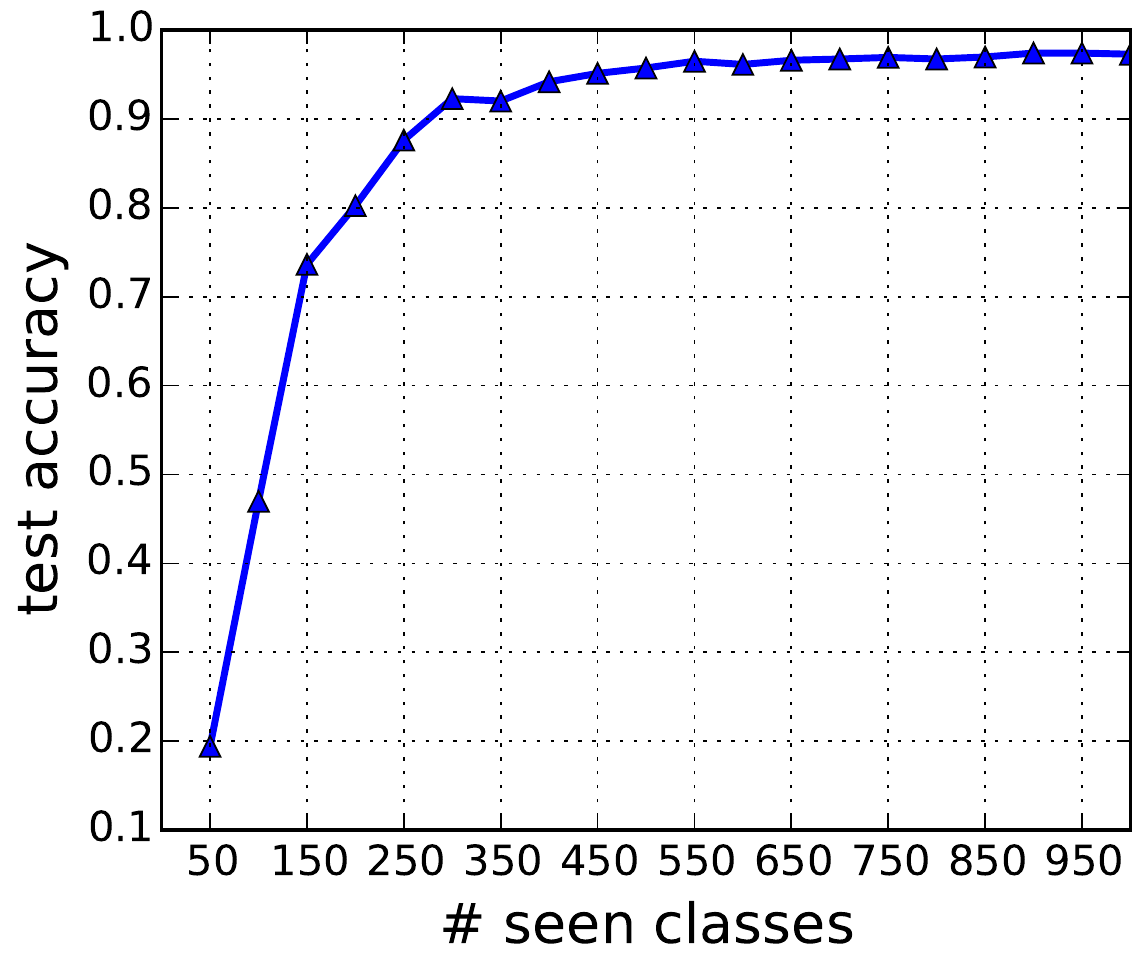}
}
\subfloat[Prediction on unseen classes]{
 \includegraphics[width=0.48\linewidth]{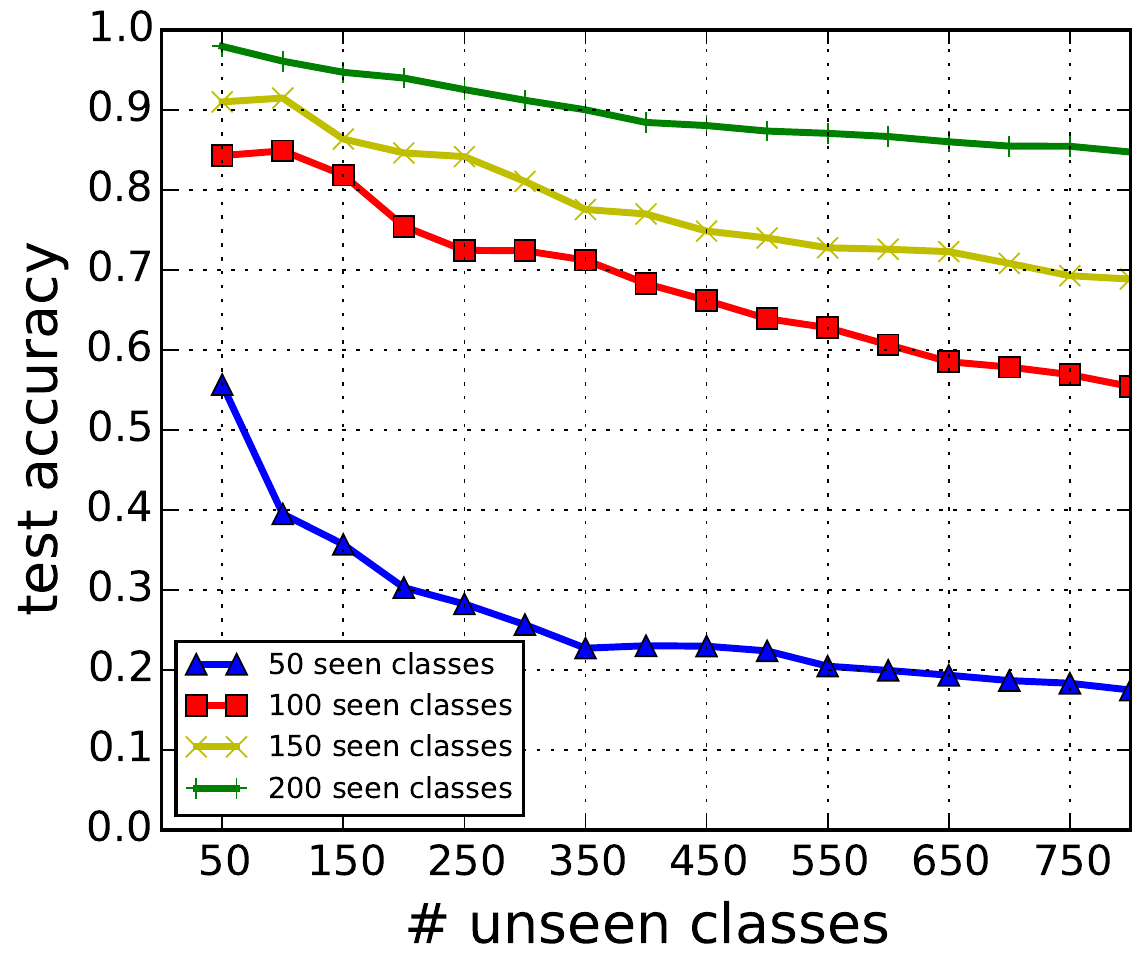}
}
\vspace{-0.6em}
\caption{Zero-shot recognition on C-MNIST by GPFR.}
\label{fig:zsl_CMNIST}
\end{figure}
% \vspace{-2.2em}
\paragraph{Experimental settings.} We use a simple CNN architecture as shared encoder whose detailed configuration is showed in Table~\ref{table:1}. We use SLP as JAFE unit where the number of neurons is 32 and the nonlinear functions are \emph{tanh}. Thus we have 3 SLP units for the whole JAFE, which are assigned to deal with digits, b-colors and f-colors, respectively. Unlike the binary selection for attributes as discussed in Sec.~\ref{chap-method-extraction}, we here use three 10-way Softmax classifiers as our attribute predictors for three attributes. Some trials indicate that the acquisition for understandings about digit is more difficult than b-color and f-color, hence we let $\alpha_{d}=1$ but $\alpha_{b}=\alpha_{f}=0.1$ in Eq.~\ref{eq:3}. For simplicity, we construct our cognitive repository without consideration about confidence margins, but instead we just select attribute feature vector into our cognitive repository which has a highest score for its attribute prediction. The whole model is optimized by Adam \cite{kingma2014adam} with mini-batch size 32.
\vspace{-1.5em}
\paragraph{Zero-shot recognition.} 
For zero-shot recognition on C-MNIST, two groups of experiments are designed to certify the capacity of GPFR on ZSL:
(1) We increase incrementally the number of seen classes from 50 to 1000 to recognize test images containing all classes; (2) We randomly select respectively 50, 100, 150 and 200 classes as seen classes to make predictions on the rest unseen classes whose number are increased incrementally from 50 to 800. For both settings, we train our JAFE for 10 epochs and then play 5 synthesizing iterations to train class predictor. For each synthesizing iteration we generate 100 pseudo representations per class (\ie pseudo size = 100) and train predictor for 1 epoch.
The results of (1) and (2) are depicted respectively in Fig.~\ref{fig:zsl_CMNIST}(a) and Fig.~\ref{fig:zsl_CMNIST}(b). As showed in Fig.~\ref{fig:zsl_CMNIST}(a), the test accuracy on all classes has reached 92.28\% just with 300 seen classes, which is improved steadily until convergence with the number of seen classes increasing. This phenomenon also can be concluded in Fig.~\ref{fig:zsl_CMNIST}(b). That is, with the number of seen classes from 50 to 200, the performance of GPFR on a certain number of unseen classes gradually become better, because the newly added classes have enriched the cognitive repository of GRFR model. However, the curves drop gradually as the number of unseen classes increasing, as showed in Fig.~\ref{fig:zsl_CMNIST}(b). For example, in the case of 200 seen classes, the GPFR performance decreased from 97.96\% to 84.75\% with unseen classes increasing from 50 to 800. To summary, the effectiveness of our GPFR is demonstrated by these experimental results.
\begin{figure}[tbp]
\begin{center}
\includegraphics[width=0.76\linewidth]{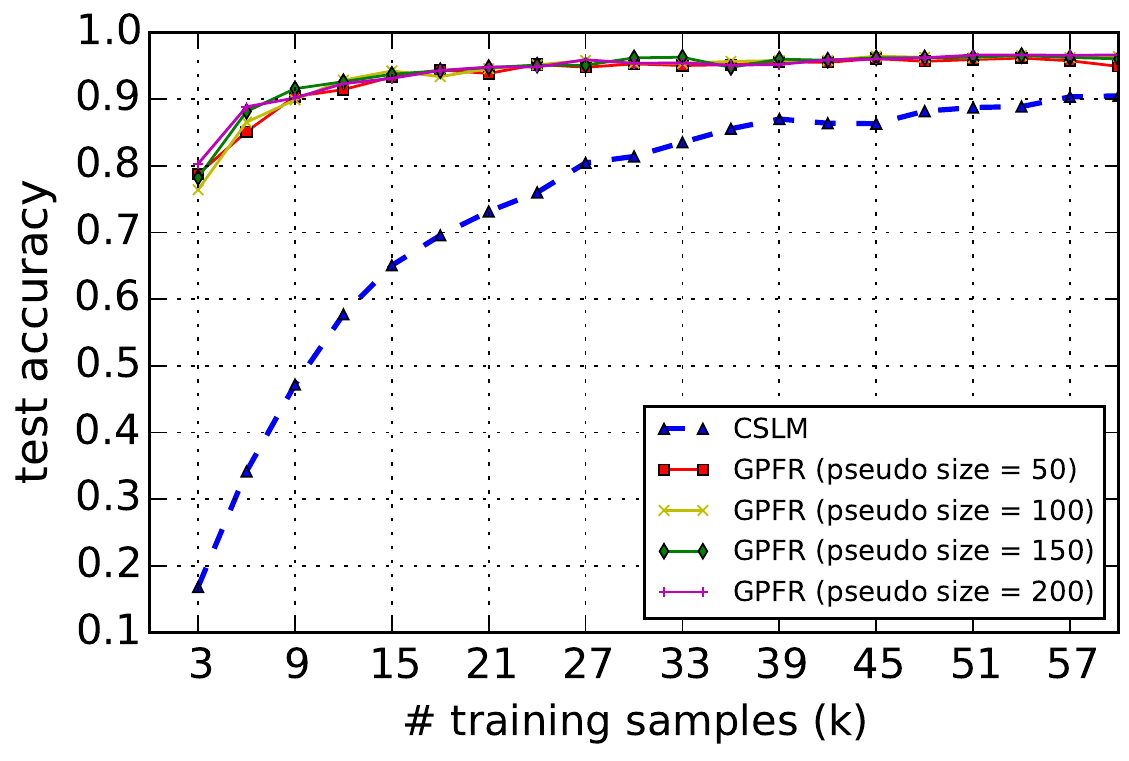}
\end{center}
\vspace{-2em}
   \caption{Fully supervised recognition comparison between our proposed GPFR and conventional supervised learning model (CSLM) on C-MNIST.}
\label{fig:supervised_CMNIST}
\end{figure}
\begin{table}[tbp]
\begin{center}
\scalebox{0.75}{
\begin{tabular}{|l|l|l|l|l|l|}
\hline
Input & Conv.1+Relu & Conv.2+Relu & Pooling & Dropout & Flatten\\
\hline
\hline
$(3, 28, 28)$ & $(32, 3, 3)$ & $(32, 3, 3)$ & $(2,2)$ & $p=0.25$ & --\\
\hline
\end{tabular}}
\end{center}
\vspace{-1.5em}
\small \caption{Configuration of our shared encoder. No padding after convolution operation and the strides are all $(1,1)$.
}
\label{table:1}
\end{table}

\begin{table*}[htpb]
\begin{center}
\scalebox{0.82}{
\begin{tabular}{|l|lllll|}
\hline
Feature & Method & aP\&Y & AwA & CUB& SUN \\
\hline
\hline
\multirow{11}{*}{\small Non-Deep}
 & Farhadi \etal \cite{farhadi2009describing} & 32.50 & --&-- & --\\
 & Mahajan \etal \cite{mahajan2011joint} & 37.93 & --&-- & --\\
 & Wang and Ji \cite{wang2013unified} & 45.05 & 42.78&-- & --\\
 & Rohrbach \etal \cite{rohrbach2013transfer} & -- &42.70&--&--\\
 & Yu \etal \cite{yu2013designing} &-- &48.30&--&--\\
 & Akata \etal \cite{akata2013label} & -- & 43.50& 18.00 & --\\
 & Fu \etal \cite{fu2014transductive} & -- & 47.10& -- & --\\
 & Mensink \etal \cite{mensink2014costa} & -- &--& 14.40 & --\\
 & Lampert \etal \cite{lampert2014attribute} & 19.10 & 40.50&-- & 52.50\\
 & Jayaraman and Grauman \cite{jayaraman2014zero} & 26.02 $\pm$ 0.05 & 43.01 $\pm$ 0.07&-- & 56.18 $\pm$ 0.27\\
 & Romera-Paredes and Torr \cite{romera2015embarrassingly} & 27.27 $\pm$ 1.62 & 49.30 $\pm$ 0.21&-- & 65.75 $\pm$ 0.51\\
\hline
\hline
\multirow{2}{*}{\small AlexNet}
 & Akata \etal \cite{akata2015evaluation} & -- &61.90& 40.30 & --\\
 & Bucher \etal \cite{bucher2016improving} & 46.14 $\pm$ 0.91 & --&41.98 $\pm$ 0.67 & 75.48 $\pm$ 0.43\\
\hline
\hline
\multirow{6}{*}{\small VGG-19}
 & Lampert \etal \cite{lampert2014attribute} & 38.16 &57.23 &-- & 72.00\\
 & Romera-Paredes \etal \cite{romera2015embarrassingly} & 24.22 $\pm$ 2.89 &75.32 $\pm$ 2.28 &-- & 82.10 $\pm$ 0.32\\
 & Zhang \etal \cite{zhang2015zero} & 46.23 $\pm$ 0.53 &76.33 $\pm$ 0.83& 30.41 $\pm$ 0.20 & 82.50 $\pm$ 1.32\\
 & Zhang \etal \cite{zhang2016zero} & 50.35 $\pm$ 2.97 &\emph{\textbf{80.46 $\pm$ 0.53}}& 42.11 $\pm$ 0.55 & 83.83 $\pm$ 0.29\\
 & Bucher \etal \cite{bucher2016improving} & 53.15 $\pm$ 0.88 &77.32 $\pm$ 1.03  &43.29 $\pm$ 0.38 & 84.41 $\pm$ 0.71\\
 & Our GPFR &\emph{\textbf{53.54 $\pm$ 2.35}}  & \emph{65.89 $\pm$ 0.52}&\emph{\textbf{44.67 $\pm$ 0.98}} &  \emph{\textbf{84.67 $\pm$ 0.47}}\\  
\hline
\end{tabular}}
\end{center}
\vspace{-1.5em}
\caption{Zero-shot recognition accuracy (\%) comparison (mean $\pm$ std) on aP\&Y, AwA, CUB-200-2011, and SUN Attribute. Here we list most popular ZSL methods with their performances, which are cited from the original papers \cite{zhang2015zero,zhang2016zero,bucher2016improving}.}
\label{table:2}
\end{table*}
\vspace{-1.25em}
\paragraph{Fully supervised recognition.} For analyzing the extensibility in supervised recognition scenario of our GPFR method, we make a comparison with conventional supervised learning model (CSLM). To explain, the so-called fully supervised recognition here is to use some training images containing 1k classes to recognize the test images which are also from these 1k classes. For GPFR, we use the same configuration with the above setting (1) except that here we own some training data for all classes. In addition to no training for JAFE and no pseudo representation synthesis, the compared model CSLM also has the same configuration with GPFR (\ie CNN+SLP+Softmax), which however adopts a naive end-to-end training strategy based on available training data for all classes. Like above experiments, for GPFR we train JAFE for 10 epochs and then played 5 synthesizing iterations, but we train CSLM for 30 epochs to get the optimum. For a more detailed comparison, we select 4 different pseudo sizes to actualize our GPFR model on this fully supervised task. We show the effect of varying number of training images on test accuracy in Fig.~\ref{fig:supervised_CMNIST}. As we see, even with few training images, our GPFR can also achieve good performance on recognition for all classes, far exceeding the results of CSLM in the same training dataset size. Moreover, no matter how much pseudo size or training dataset size is, the results of GPFR are always superior to CSLM. Therefore, we believe that our GPFR realizes a feature-level data augmentation for supervised learning and the synthetic pseudo representations fill the gaps between training domain and test domain.

%-------------------------------------------------------------------------
\subsection{Benchmark Comparison} \label{chap-expre-benchmark}
We test out method on other four benchmark image datasets for ZSL, \ie aP\&Y, AwA, CUB-200-2011 and SUN Attribute. The statistics of each dataset have been summarized in Table~\ref{table:3}. More specifically, each image of aP\&Y, CUB-200-2011 and SUN Attribute has its own image-level attribute notations. But for AwA, the images from one same class share the same class-level attribute notations, which are loosely regarded as image-level ones for all images from that same class. For seen classes we utilize their image-level attribute notations to train JAFE, but for unseen classes we take the means of their image-level attribute notations as their class-level attribute descriptions. To make comparisons with previous works, we use the same seen/unseen splits as \cite{farhadi2009describing} (aP\&Y), \cite{akata2015evaluation} (CUB-200-2011), \cite{lampert2009learning} (AwA) and \cite{jayaraman2014zero} (SUN Attribute). 
\begin{table}[htbp]
\begin{center}
\scalebox{0.77}{
\begin{tabular}{|l|llc|}
\hline
Dateset & \# Images & \# Attributes & \# Seen/Unseen classes\\
\hline
\hline
aP\&Y & 15,339 & 64 (image-level) & 20 / 12 \\
AwA & 30,475 & 85 (class-level)& 40 / 10\\
CUB-200-2011 & 11,788 & 312 (class-level)& 150 / 50 \\
SUN Attribute & 14,340 & 102 (class-level)& 707 / 10 \\
\hline
\end{tabular}}
\end{center}
\vspace{-1.5em}
\caption{Statistics of the four benchmark dataset.}
\label{table:3}
\end{table}

\begin{figure*}[tbp]
\centering
\subfloat[aP\&Y]{
 \includegraphics[width=0.397\linewidth]{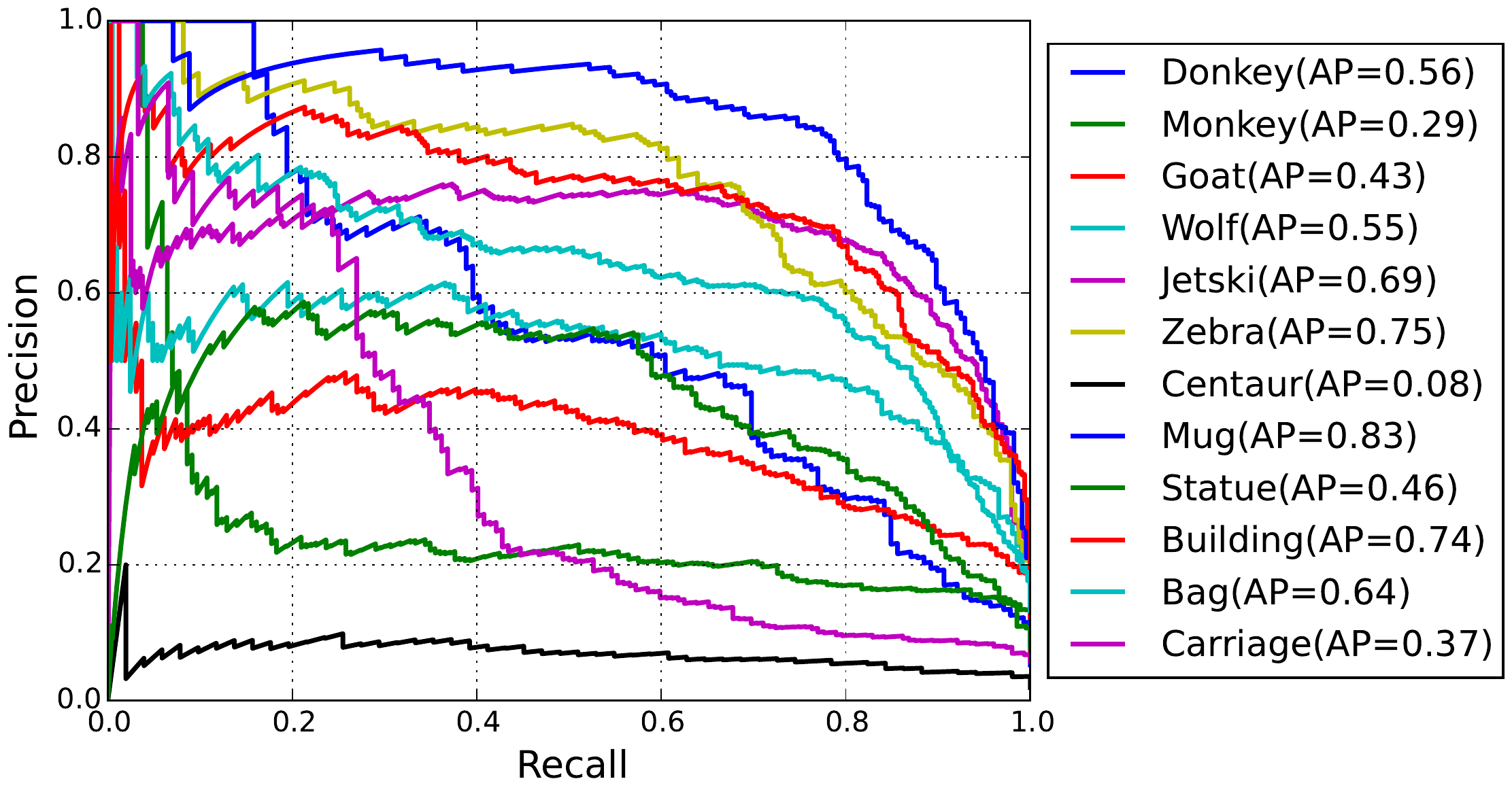}
}
\hspace{5.6em}
\subfloat[AwA]{
 \includegraphics[width=0.428\linewidth]{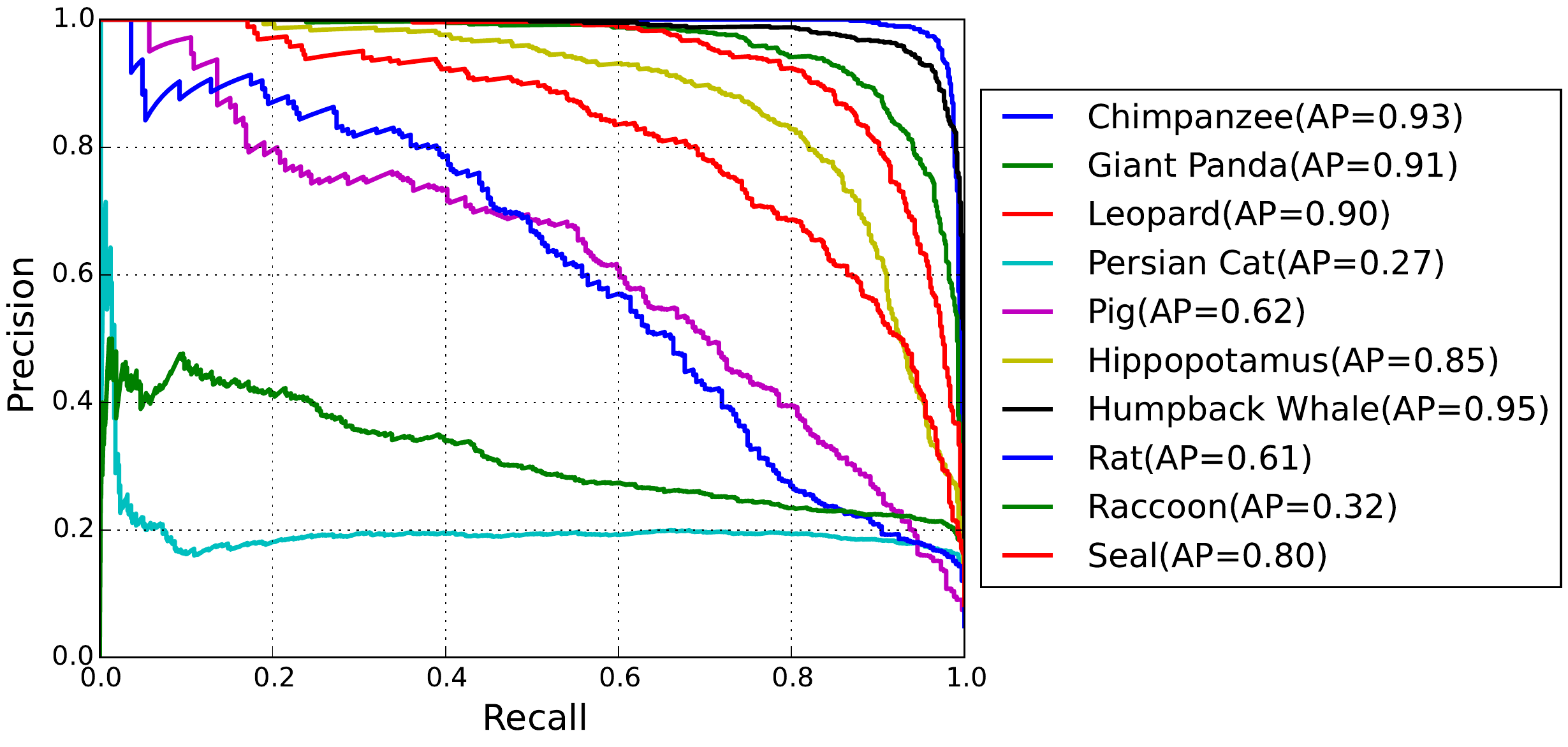}
}
\vspace{-0.7em}
\subfloat[CUB-200-2011]{
 \includegraphics[width=0.455\linewidth]{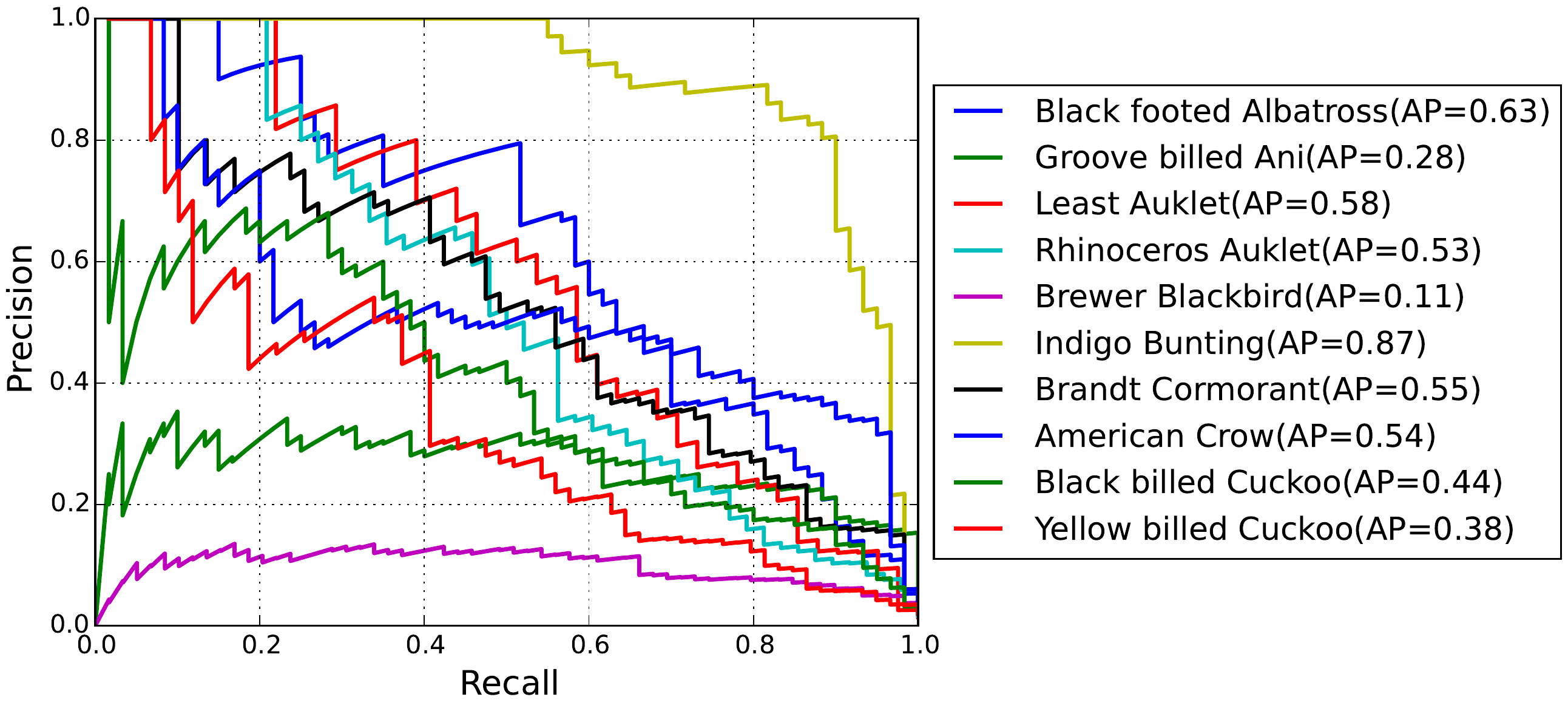}
}
\hspace{2.7em}
\subfloat[SUN Attribute]{
 \includegraphics[width=0.425\linewidth]{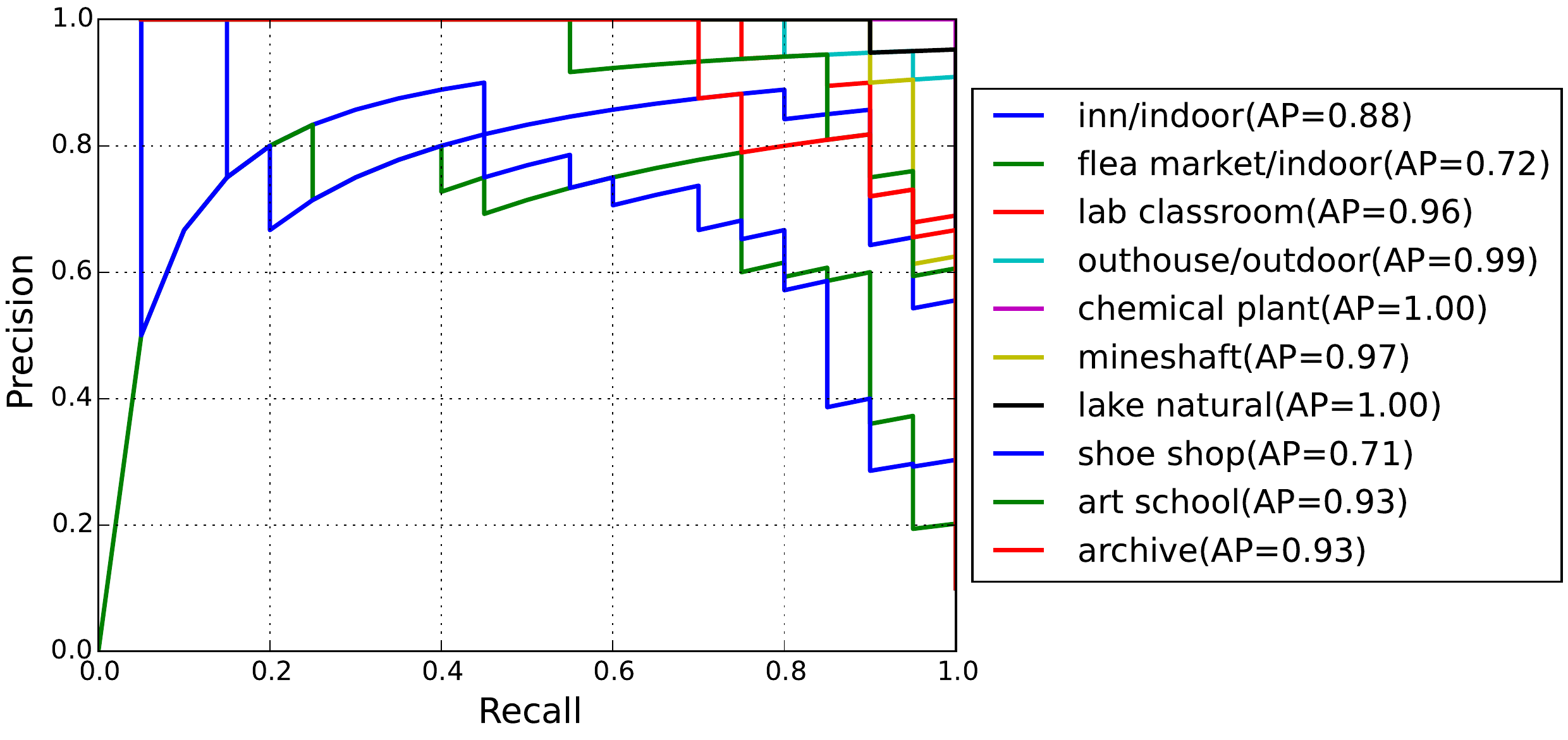}
}
\vspace{-0.6em}
\caption{Precision-Recall curves for unseen classes in four benchmark datasets. For CUB-200-2011, we show the first 10 classes from 50 test classes. Best viewed in color.}
\label{fig:ap_curves}
\end{figure*}
\vspace{-0.8em}
\paragraph{Experimental settings.} Like previous work \cite{zhang2015zero,zhang2016zero,bucher2016improving}, we utilize Keras with ``VGG19'' pre-trained model \cite{simonyan2014very} without fine-tuning on these benchmark datasets to extract the 4096-dimensional CNN feature vector on the first fully connected layer for each image. Moreover, we randomly select respectively 3, 3, 10 and 30 classes from the seen classes set of aP\&Y, AwA, CUB-200-2011 and SUN Attribute to validate our GPFR model. It is worth mentioning that the network configuration and the hyperparameter settings\footnote{Detailed settings will be listed in our supplementary material.} for different benchmarks are somewhat different, but their JAFE units are all two-layer MLPs. For aP\&Y, AwA, CUB-200-2011 and SUN Attribute, the number of neurons for each two-layer JAFE unit are 64-12, 32-6, 64-6 and 48-6, respectively. The class predictors for different datasets are all Softmax classifiers, which are designed to match their corresponding numbers of target classes, \ie 12+3, 10+3, 50+10 and 10+30 respectively. The whole model is optimized by Adam \cite{kingma2014adam} or RMSprop \cite{tieleman2012lecture} with mini-batch size 256 for JAFE but 64 for class predictor.
\vspace{-1.0em}
\paragraph{Zero-shot recognition.} The task of zero-shot recognition consists in identifying for a novel image which of unseen classes it belongs to. We summarize our comparison with state-of-the-art ZSL methods in Table~\ref{table:2}, where the results of previous methods on four benchmark datasets have been listed and the `--' indicates that no experiments have been performed on this dataset in original paper. For comparison we also reported average results over 3 trials. As we see, our proposed method outperforms slightly state-of-the-art performance in aP\&Y, CUB-200-2011 and SUN Attribute. Note that a gap appears between our result and the state-of-the-art on AwA. This can be explained by the fact that the rough class-level attribute notations may not be consistent with all real images from this same class. Naturally, using the rough class-level attribute notations as image-level ones for all images from the same class cannot result in a good JAFE module for GPFR.
\vspace{-1em}
\paragraph{Zero-shot retrieval.} The task of zero-shot retrieval is to search some images related to the specified attribute descriptions of unseen classes. Here we use above well-trained GPFR model to rank all test images along each unseen class based on their final Softmax scores output by the class predictor. Table~\ref{table:4} presents the comparative results for mAP in four benchmark datasets. Note that our GPFR significantly and consistently outperforms state-of-the-art ZSL methods by 14.82\%, 3.37\%, 21.35\% and 10.73\% on four benchmark datasets respectively and 12.68\% on average, which are tremendous improvements for zero-shot retrieval tasks. We believe it benefits from the good tradeoff of our GPFR model in class prediction, namely one misclassified image still owns high score for its correct class. Fig.~\ref{fig:ap_curves} shows precision-recall curves and AP values for all unseen classes of four benchmark datasets. Compared with the curves in \cite{bucher2016improving}, 
\begin{table}[tbp]
\begin{center}
\scalebox{0.85}{
\begin{tabular}{|l|llll|l|}
\hline
Method & aP\&Y & AwA & CUB & SUN & Ave.\\
\hline
\hline
Zhang \etal \cite{zhang2015zero} &15.43&46.25& 4.69 & 58.94 & 31.33\\
Zhang \etal \cite{zhang2016zero} &38.30& 67.66& 29.15 & 80.01&  53.78\\
Bucher \etal \cite{bucher2016improving} &36.92& 68.10& 25.33  & 52.68&45.76 \\
Our GPFR  & \emph{\textbf{53.12}} & \emph{\textbf{71.47}}& \emph{\textbf{50.50}} & \emph{\textbf{90.74}}& \emph{\textbf{66.46}} \\
\hline
\end{tabular}}
\end{center}
\vspace{-1.5em}
\caption{Zero-shot retrieval mAP (\%) comparison on four benchmark datasets. The results of other methods are cited from \cite{zhang2016zero,bucher2016improving}.}
\label{table:4}
\end{table}
\vspace{-0.1em}
\begin{figure}[tbp]
\begin{center}
\includegraphics[width=1.0\linewidth]{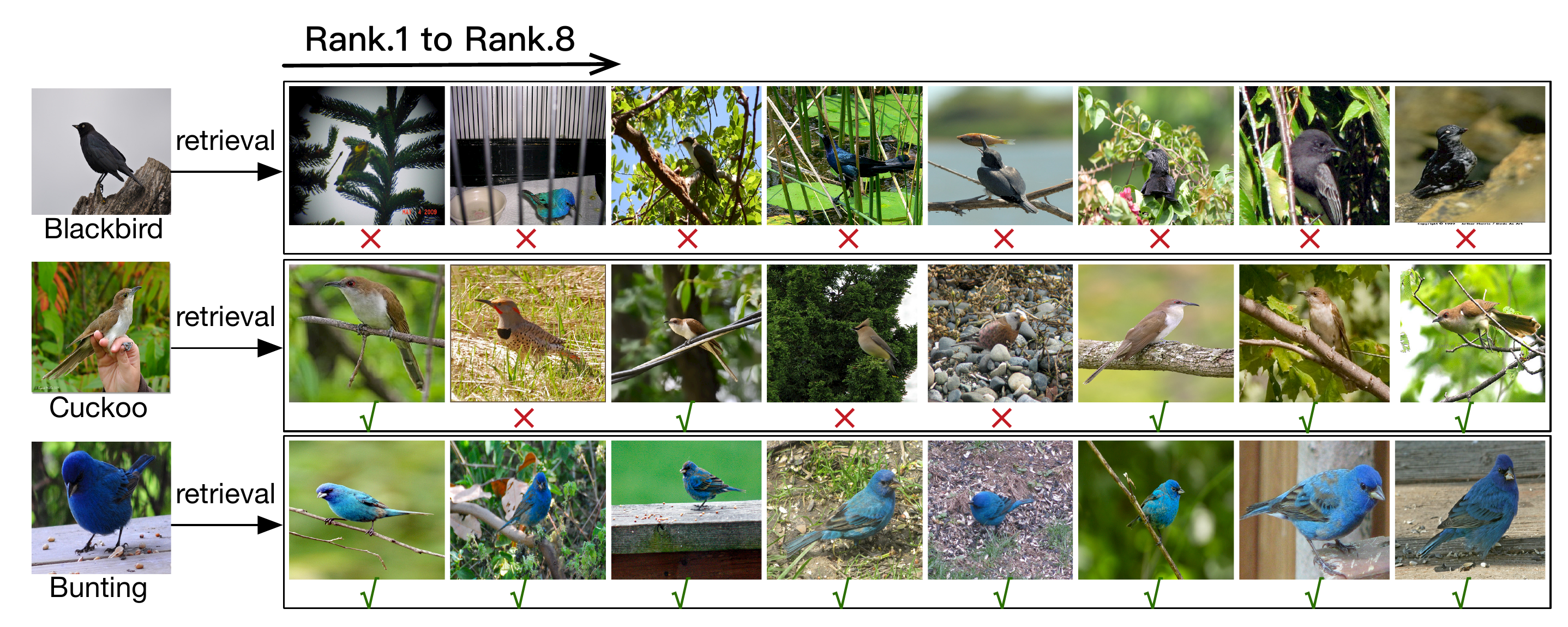}
\end{center}
\vspace{-1.9em}
  \caption{Top-8 zero-shot retrieval results by our GPFR for class \emph{``Brewer Blackbird''}, \emph{``Black billed Cuckoo''} and \emph{``Indigo Bunting''} (from top to down) in CUB-200-2011.}
\label{fig:cub_retrieval}
\end{figure}
our method obviously has a larger area under the curves for arbitrary dataset. For intuition, we select 3 unseen classes of CUB-200-2011 in different difficulty levels to visualize our zero-shot retrieval results in Fig.~\ref{fig:cub_retrieval} with top-8 returns. Two most frequent failure modalities are: (1) retrieved images contains cluttered background, and (2) a visually similar category is confused with the target category. These failure modalities are still some difficult challenges for ZSL as well as fine-grained recognition.

\subsection{Visualization for Pseudo Representations} \label{chap-expre-visual}
\begin{figure}[htbp]
\begin{center}
\includegraphics[width=1\linewidth]{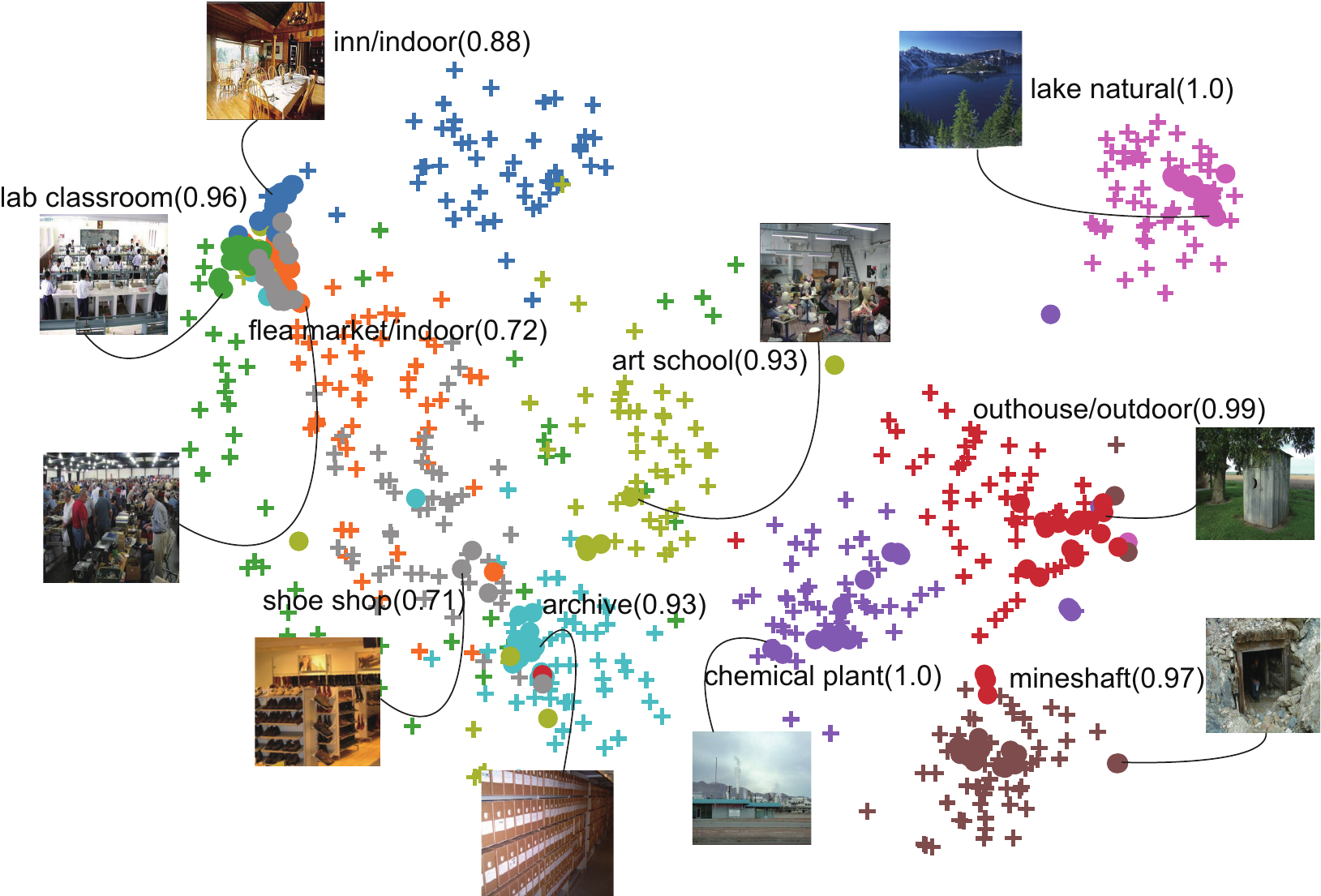}
\end{center}
\vspace{-1.5em}
   \caption{t-SNE visualization for real and pseudo feature representations of 10 unseen classes in SUN Attribute. The 10 unseen classes are indicated with 10 different colors respectively. Solid points denote the real feature representations but plus signs denote the pseudo ones. The number in brackets is the retrieval AP (\%). Best viewed in color.}
\label{fig:visualization}
\end{figure}

For better comprehension of our GPFR method, we visualize the pseudo feature representations as well as the real feature representations extracted by JAFE on SUN Attribute using t-SNE \cite{maaten2008visualizing} in Fig.~\ref{fig:visualization}. As we can see, the real feature representations and the pseudo ones gather together nicely for some unseen classes that owns distinguishing characteristics compared to other classes, \eg \emph{``lake natural''}, \emph{``mineshaft''} and \emph{``chemical plant''}, which facilitates a high retrieval AP reasonably. However, for those classes that owns some common characteristics, \eg \emph{``inn/indoor''}, \emph{``lab classroom''}, \emph{``flea market/indoor''} and \emph{``shoe shop''}, it is difficult to distinguish them by real feature representations since they are all indoor scenes. Hence, their gathering of pseudo and real feature representations are relatively poor, naturally leading to a relatively low retrieval AP.

%------------------------------------------------------------------------
\vspace{-0.8em}
\section{Related Work}\label{chap-relatedwork}
\paragraph{Two-stage probability based methods.} 
Some ZSL methods are based on a two-stage recognition: attribute prediction and classification inference.
\cite{farhadi2009describing,lampert2009learning,palatucci2009zero,yu2013designing,lampert2014attribute} regarded 
each unseen class as a binary vector, \ie signature, where each entry delegates the presence or absence 
of one attribute. For a new image, its attributes are first predicted, and then it will be mapped to an unseen 
class whose signature is most similar to the predicted attributes. 
Explicitly, \cite{lampert2009learning,lampert2014attribute} proposed a probabilistic Directed Attribute Prediction (DAP) framework where attribute probabilistic classifiers are learned firstly and then a MAP step is performed for
seeking the most promising unseen class.
\cite{yu2010attribute} proposed an Author-Topic model to describe the attribute-specified distributions of image features.
\cite{wang2013unified} constructed a Bayesian Network (BN) based unified model to capture the object-dependent and object-independent attribute relationships.
\cite{suzuki2014transfer} proposed a weighted version of DAP based on observation probability of attributes. 
\cite{hariharan2012efficient} developed a max-margin multi-label classification formulation (M3L) for attribute 
prediction.
\vspace{-2em}
\paragraph{Semantic embedding based methods.}
Other ZSL methods are based on view of embedding.
Their key idea is to encode both images and labels into a common space and then learn a discriminative compatibility/similarity function, where image embedding can be feature vector
or other related representations, and label embedding can be particular coding, \eg attributes, or available text corpus, \eg English Wikipedia.
\cite{akata2013label,romera2015embarrassingly} utilized label embedding to construct a 
common space where the compatibility between images and labels can be measured.
\cite{akata2015evaluation,frome2013devise,lei2015predicting} leveraged CNN based features \cite{krizhevsky2012imagenet,szegedy2015going} as image embedding to learn compatibility function.
\cite{fu2014transductive,kodirov2015unsupervised} built better embedding by alleviate domain shift problem
which was considered to exist between source domain (seen classes) and target domain (unseen classes).
\cite{zhang2015zero} embedded source and target domain data into semantic space, \ie mixture proportions of seen classes.
\cite{xian2016latent} learned a collection of linear models to construct overall nonlinear latent embedding models.
\cite{zhang2016zero} presented a general probabilistic model embedding both domains to a joint latent space.
\cite{bucher2016improving} suggested to better control semantic embedding of images by 
metric learning \cite{mensink2012metric}. 
Such works essentially translated zero-shot tasks into sub-task associations of semantic embedding and similarity measurement, exhibiting excessive reliance on the capability of common embedding spaces.

Moreover, few related approaches follow different perspectives, such as semantic transfer \cite{rohrbach2013transfer}, co-occurrence statistics of visual concepts \cite{mensink2014costa}, random forest \cite{jayaraman2014zero} and semantic manifold structure \cite{fu2015zero}. Different from these aforementioned methods, our method propose to generate pseudo feature representations to enrich our comprehension for novelties, showing good extensibility in conventional supervised visual recognition. 

%------------------------------------------------------------------------
\section{Conclusion}\label{chap-conclusion}
In this paper we develop a novel ZSL framework by generating pseudo feature representations (GPFR). We summarize all accessible understandings related to attributes to construct our surmises for unseen classes. Compared with most existing ZSL methods, the superiorities of our method are the simplicity for implementation and the extensibility for supervised recognition scene. Our method on four benchmark datasets shows compelling performance for zero-shot recognition task, and leads to significant improvement to state-of-the-art mAP for zero-shot retrieval task.

{\small
\bibliographystyle{ieee}
\bibliography{ref}
}

\end{document}